\documentclass[10pt,twocolumn,letterpaper]{article}

\usepackage{authblk}
\usepackage{cvpr}              % To produce the CAMERA-READY version
\usepackage{times}
\usepackage{epsfig}
\usepackage{graphicx}
\usepackage{amsmath}
\usepackage{amssymb}
\usepackage{multirow}
\usepackage{caption}
\usepackage{subcaption}
\usepackage{booktabs}
\usepackage[normalem]{ulem}
\usepackage{bm}
\usepackage{xcolor}

\usepackage[pagebackref,breaklinks,colorlinks]{hyperref}

\usepackage[capitalize]{cleveref}
\crefname{section}{Sec.}{Secs.}
\Crefname{section}{Section}{Sections}
\Crefname{table}{Table}{Tables}
\crefname{table}{Tab.}{Tabs.}

\def\cvprPaperID{27} % *** Enter the CVPR Paper ID here
\def\confName{NTIRE}
\def\confYear{2022}

\newcommand\norm[1]{\left\lVert#1\right\rVert}

\begin{document}

%%%%%%%%% TITLE
% \title{Image Compression for Face Recognition with Identity Preserving Reconstruction Loss}
% \title{Learned Image Compression for faces}
% \title{Learned Image Compression for face recognition}
% \title{Identity Preserving Learned Image Compression}
\title{Identity Preserving Loss for Learned Image Compression}

% 11/1/2021 updated

\author[1,2]{Jiuhong Xiao \thanks{Work done during an internship at Amazon.}}
\author[2]{Lavisha Aggarwal}
\author[2]{Prithviraj Banerjee}
\author[2]{Manoj Aggarwal}
\author[2]{Gerard Medioni}
{
\makeatletter
\renewcommand\AB@affilsepx{\protect\quad\protect\Affilfont}
\makeatother
\affil[1]{New York University\quad}
}
\affil[2]{Amazon}

\affil[ ]{\tt\small {jx1190@nyu.edu\quad aggarl@amazon.com\quad banerjep@amazon.com\quad manojagg@amazon.com\quad medioni@amazon.com}}

\makeatletter
\def\@maketitle{%
\newpage%
\null%
\vskip 0em % <================================================ space c)
\begin{center}%
    \let\footnote\thanks %
    {\LARGE \@title %
      \par % <================================================= space d)
    }
%   \vskip 1.5em % <=========================================== space d)
    {\large
     \lineskip 0em
     \begin{tabular}[t]{c}
        \baselineskip=16pt
        \@author
     \end{tabular}
     \par% <=================================================== space e')
    }
\end{center}
\par % <======================================================= space f)
\vskip 1.5em} % <============================================== space f)
\makeatother

\maketitle

%%%%%%%%% ABSTRACT
\begin{abstract} % 11/1/2021 updated
Deep learning model inference on embedded devices is challenging due to the limited availability of computation resources. A popular alternative is to perform model inference on the cloud, which requires transmitting images from the embedded device to the cloud. Image compression techniques are commonly employed in such cloud-based architectures to reduce transmission latency over low bandwidth networks. This work proposes an end-to-end image compression framework that learns domain-specific features to achieve higher compression ratios than standard HEVC/JPEG compression techniques while maintaining accuracy on downstream tasks (e.g., recognition). Our framework does not require fine-tuning of the downstream task, which allows us to drop-in any off-the-shelf downstream task model without retraining. We choose faces as an application domain due to the ready availability of datasets and off-the-shelf recognition models as representative downstream tasks. We present a novel Identity Preserving Reconstruction (IPR) loss function which achieves Bits-Per-Pixel (BPP) values that are $\sim38\%$ and $\sim42\%$ of CRF-23 HEVC compression for LFW (low-resolution) and CelebA-HQ  (high-resolution) datasets, respectively, while maintaining parity in recognition accuracy. The superior compression ratio is achieved as the model learns to retain the domain-specific features (e.g., facial features) while sacrificing details in the background. Furthermore, images reconstructed by our proposed compression model are robust to changes in downstream model architectures. We show at-par recognition performance on the LFW dataset with an unseen recognition model while retaining a lower BPP value of \text{$\sim38\%$} of CRF-23 HEVC compression.

\end{abstract}

%%%%%%%%% BODY TEXT
\section{Introduction}

\begin{figure}[!htb]
    \centering
    \includegraphics[scale=0.40]{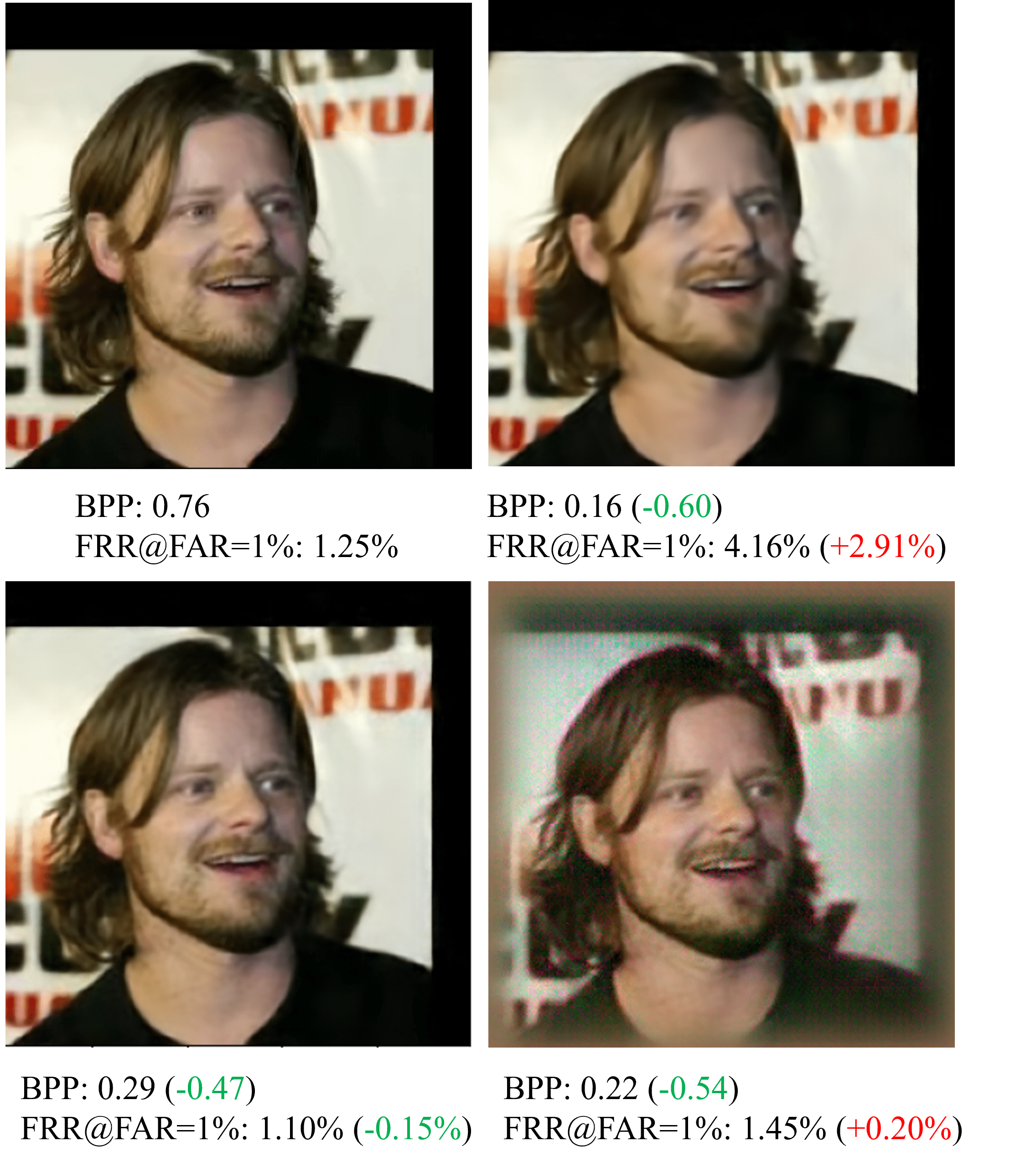}
    \caption{Identity preserving loss for learned image compression. Top left: HEVC baseline; Top right: The result shows that if the compression model only optimizes rate-distortion trade-off, the significant distortion appears on the face region and degrades recognition performance; Bottom left: The compression model optimizing Identity Preserving Reconstruction (IPR) loss can maintain image quality with improved recognition accuracy; Bottom right: If the compression model optimizes Identity Preserving (IP) loss, it sacrifices the image quality of background and maintains the clear face region with slightly lower accuracy and much lower BPP than baseline.}
    \label{fig:teaser}
\end{figure}

There is an increasing trend to perform tasks like object detection and semantic segmentation on mobile or resource-constrained local devices. However, at-edge computing often cannot afford the high computation complexity of large CNN or Transformer models. Instead of reducing model complexity on local devices, a popular solution is to transfer images from the device to a remote server and execute computational-intensive tasks on the cloud.

This paper focuses on an end-to-end image compression framework for the face recognition domain due to the ready availability of datasets, and off-the-shelf recognition models as representative downstream tasks. For face-related tasks, recent works focus on high-resolution face images for high-precision face recognition \cite{high-face-recognition} and 3D face reconstruction \cite{high-3d-recon}. Local devices with limited resources will show high inference latency on these tasks. There is also a potential requirement to store raw images on database servers for the downstream pipeline. Under these circumstances, a remote server needs to receive and store high-resolution images. Image compression methods help reduce the transmission latency and storage space on the server.

Standard lossy compression techniques like JPEG \cite{jpeg} and HEVC \cite{hevc} reduce image file size without noticeable distortion on images. With the growing popularity of deep learning \cite{lecun2015deep}, an increasing number of recently published works are using CNN models to compress images to achieve a higher compression ratio and less notable compression artifacts \cite{variational,HIFIC,Agustsson2017SofttoHardVQ,rnn}. Further works \cite{Chamain2021EndtoEndOI, chamain2021end} indicate that end-to-end image compression training with downstream tasks (like object detection) can highlight regions relevant to the downstream task, and further increase the compression ratio with improved downstream performance.

Since our compression model only receives face images, it allows us to tune the compression model against the face domain and further increase the compression ratio. We regard our compression model as a plug-in module between the image capture module and the downstream system, so we update only the compression model during training. We use HEVC(CRF=23) and JPEG as standard compression baselines, and provide a performance comparison for a) low-resolution vs. high-resolution, and  b) unaligned vs. aligned face datasets.

Our method has multiple advantages: First, our method does not require additional retraining or fine-tuning of the downstream recognition model. Second, we show that the compression models trained for a specific domain (faces in our case) show a higher compression ratio and recognition accuracy than those trained with generic images. Third, a new framework with Identity Preserving Reconstruction (IPR) loss or Identity Preserving (IP) loss is proposed to improve the recognition performance of the images generated from the compression model. The results show that all models with IPR loss have improved recognition performance for the low-resolution dataset. Our method achieves a high compression ratio with slightly lower recognition accuracy for the high-resolution dataset. Lastly, we present an evaluation of our proposed algorithm's robustness to downstream model changes, and show improved recognition accuracy and high compression ratio with an unseen recognition model. 
% Finally, our framework shows the robustness to the change of downstream model by performing improved recognition accuracy and high compression ratio with an unseen recognition model. 
%-------------------------------------------------------------------------
% TODO: Change the framework fig
\begin{figure*}[!htb]
    \centering
    \includegraphics[scale=0.55]{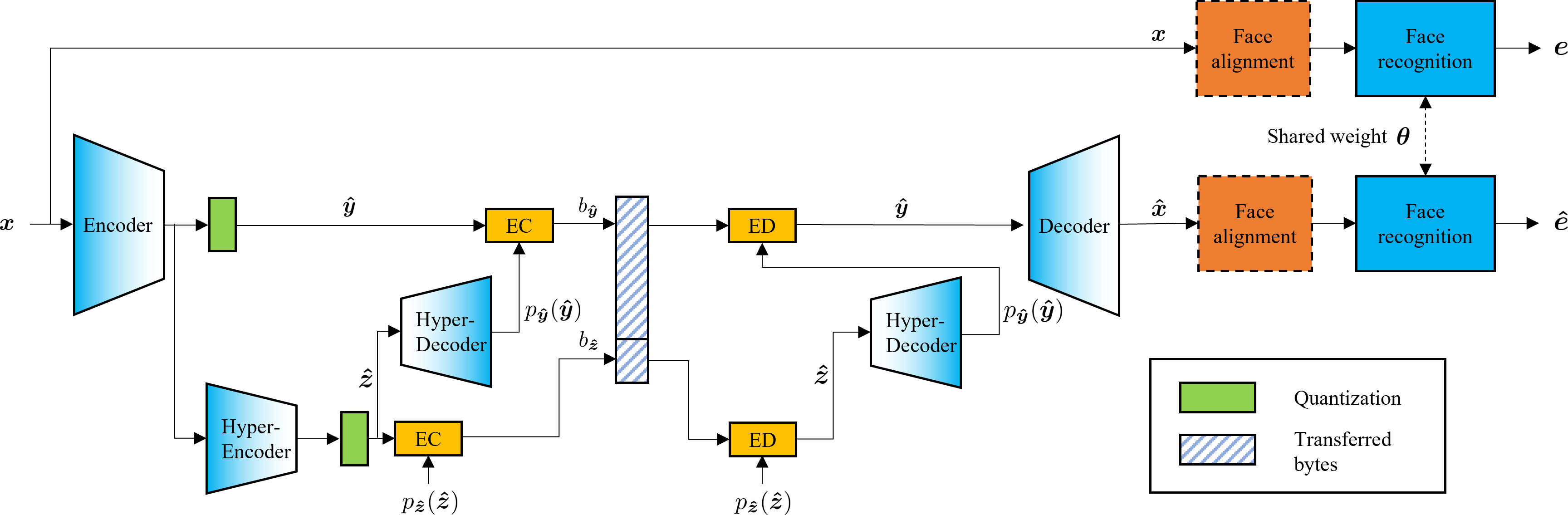} 
    \caption{Framework for the compression model with face recognition system. The face alignment is optional depending on if the images are pre-aligned.}
    \label{fig:framework}
\end{figure*}

\section{Related Work}

For lossy compression, a learnable image compression method should consider both the network backbone and the entropy model \cite{survey}. There are two approaches for the network backbone: autoencoder-like architectures \cite{variational,johnston2019computationally,HIFIC} and recurrent frameworks \cite{rnn}. Autoencoder-like architectures can reduce dimensionality to produce a compact latent representation. Recurrent frameworks are used to encode the residue to compress the image progressively.

For the entropy model, there are several options like context-based prediction \cite{NEURIPS2018_53edebc5,variational} or calculating histogram of latent representation \cite{Theis2017LossyIC}. One of the most widely-used frameworks, the hyper-prior method \cite{variational}, follows the VAE architecture and uses the hyper-prior model as an entropy model. From the survey \cite{survey}, the authors observe that the hyper-prior \cite{variational} method outperforms other methods for its outstanding entropy model.

There are multiple variants of the hyper-prior method: \cite{johnston2019computationally} conducts network architecture search (NAS) on the hyper-prior model to reduce the amount of computation, while not reducing the encoded file size. Alternately, \cite{HIFIC} integrates GAN with the hyper-prior model to generate images that achieve a high subjective image quality (realistic texture). However, such a method encourages ``photo-realism'' over preserving the fine information detail in the images, which may impact downstream tasks that need original details (such as recognition algorithms).

% Considering the variants of the hyper-prior method, High-Fidelity Generative Image Compression \cite{HIFIC} uses GAN with the hyper-prior model to improve high-fidelity images and reaches state-of-the-art to reconstruct realistic images, while we find that even though the output images have high subjective image quality (realistic texture), they have high distortion error, which means they are not matching the original images. Another variant \cite{johnston2019computationally} focuses on doing NAS on the hyper-prior method to reduce the amount of computation while not reducing the encoded file size.

For task-specific image compression, the work \cite{Chamain2021EndtoEndOI,chamain2021end} shows that by jointly optimizing the compression model and an object detection model, the compressed images highlight the task-related region and do not decrease the downstream performance significantly. Some works \cite{jpeg1,jpeg2} also perform task-based JPEG compression. Recent work \cite{torfason2018towards} points out that task-specific compression can be done without decoding for classification and semantic segmentation tasks by jointly training both the decoder and the classifier using intermediate code. Compared to these works, our method applies the task-specific image compression method to the face imagery domain, without any requirements on optimizing/fine-tuning the downstream recognition model, and the encoder-decoder-recognition paradigm instead of the encoder-recognition \cite{torfason2018towards} paradigm for increased generalizability to different downstream recognition models.

For face-specific learnable compression, \cite{wang2019scalable} uses deep-feature-based compression. It applies compression to the feature map extracted by FaceNet \cite{Schroff2015FaceNetAU}. Deep-feature-based compression only allows compression for aligned images and depends on the FaceNet feature tuned to face recognition. \cite{wang2021one} uses a talking-head video synthesis model to allow face-specific video compression. Our method focuses on both unaligned and aligned still images and can be generalized to other tasks than recognition. Besides, \cite{wang2019scalable} uses an unlearned entropy model, while our method uses a learned entropy model, which allows a higher compression ratio. Overall, our method can show higher flexibility, and compression performance than \cite{wang2019scalable}.

\section{Compression for Face Recognition}

Our proposed framework is shown in Fig. \ref{fig:framework}. Images $\bm x$ are first encoded and quantized to discrete latent representation $\bm{\hat y}$. We follow the hyper-prior method \cite{variational} to set the entropy model. The hyper encoder-decoder outputs discrete hyper-latent representation $\bm{\hat z}$ and predicts the distribution of $\bm{\hat y}$ using the entropy model $ p_{\bm{\hat y}}(\bm{\hat y})$. An entropy encoder (EC) and an entropy decoder (ED) are for lossless compression. For EC, $p_{\bm{\hat y}}(\bm{\hat y})$ is used to encode $\bm{\hat y}$ to transferable bytes $b_{\bm{\hat y}}$ and a separate entropy model $p_{\bm{\hat z}}(\bm{\hat z})$ with a factorized prior distribution is used to encode $\bm{\hat z}$ to $b_{\bm{\hat z}}$. $b_{\bm{\hat y}}$ and $b_{\bm{\hat z}}$ make up the payload that needs transmission over the network. After transmission, ED will transform those bytes back to discrete latent and hyper-latent representation. Finally, the decoder reconstructs the images $\hat{\bm{x}}$.

The existing compression methods optimize reconstruction loss to maintain image quality and bit-rate loss to reduce the file size:
\begin{equation}
\mathcal{L}_\textrm{R}=\mathcal{L}_\textrm{rec}+\lambda_\textrm{rate}\mathcal{L}_\textrm{rate}
\label{equ:rd}
\end{equation}
\begin{equation}
    \mathcal{L}_\textrm{rec} =  \mathbb{E}_{\bm x \sim p_{\bm{x}}}[d(\bm{x}, \bm{\hat x})]
\label{equ:d}
\end{equation}
\begin{equation}
    \mathcal{L}_\textrm{rate} = \mathbb{E}_{\bm{x} \sim p_{\bm{x}}}[-\log(p_{\bm{\hat y}}(\bm{\hat y}))-\log(p_{\bm{\hat z}}(\bm{\hat z}))]
\end{equation}
where $\mathcal{L}_\textrm{rec}$ is the reconstruction loss, $\mathcal{L}_\textrm{rate}$ is the bit-rate loss, $d(\bm{x}, \bm{\hat x})$ is the distortion function, and $\lambda_\textrm{rate}$ is the weight for bit-rate loss. We name the total loss function $\mathcal{L}_\textrm{R}$ as \textbf{Reconstruction-only} loss. When selecting the MSE function as the reconstruction loss, $d(\bm{x}, \bm{\hat x})$ is defined as:
\begin{equation}
    d(\bm{x},\bm{\hat x}) = \norm{\bm{x} - \bm{\hat x}}^2
\end{equation}

Other options are the MS-SSIM function \cite{1292216} or some perceptual metrics that can measure the distortion between original images $\bm{x}$ and reconstructed images $\bm{\hat x}$.

To make quantization differentiable in training, the hyper-prior method \cite{variational} adds uniform noise to simulate quantization error and calculate $\mathcal{L}_\textrm{rate}$. However, in \cite{Theis2017LossyIC}, using additive noise as the quantization alternative adds strong visual artifacts into decompressed images for JPEG compression. Therefore, we still use (hyper-)latent variables ($\bm{\hat y}$ and $\bm{\hat z}$) with additive noise when calculating $p_{\bm{\hat y}}(\bm{\hat y})$, $p_{\bm{\hat z}}(\bm{\hat z})$ for $\mathcal{L}_\textrm{rate}$, but send the (hyper-)latent variables quantized by rounding function (used in the actual encoding process) to (hyper-)decoder, using a differentiable  back-propagation function like \cite{HIFIC}.

\subsection{Identity Preserving Reconstruction Loss}
Our goal is to improve the compression ratio without degrading recognition performance. For this task, we use the CurricularFace \cite{huang2020curricularface} pre-trained model. We do not fine-tune the model weights (we freeze the weights instead).

To make the compression model compatible with the recognition model, we derive the \textbf{Identity Preserving (IP)} loss function, including the bit-rate loss and identity loss to reduce the distance between $\bm{e}$ and $\bm{\hat e}$ (Fig. \ref{fig:framework}) for identity recognition with smaller file size:

\begin{equation}
\mathcal{L}_\textrm{IP}= \lambda_\textrm{rate}\mathcal{L}_\textrm{rate}+\lambda_\textrm{id}\mathcal{L}_\textrm{id}
\label{equ:re}
\end{equation}
\begin{equation}
\mathcal{L}_\textrm{id}= \mathbb{E}_{\bm{x} \sim p_{\bm{x}}} \left [1 - \frac{\bm{e} \cdot \bm{\hat e}}{\norm{\bm{e}}\norm{\bm{\hat e}}} \right ]
\label{equ:id}
\end{equation}
where $\mathcal{L}_\textrm{id}$ is the identity loss. Since $\mathcal{L}_\textrm{IP}$ only optimizes recognition performance, we further derive \textbf{Identity Preserving Reconstruction} (IPR) loss function to jointly optimize image quality and recognition performance:

\begin{equation}
\mathcal{L}_\textrm{IPR}= \mathcal{L}_\textrm{rec}+\lambda_\textrm{rate}\mathcal{L}_\textrm{rate}+\lambda_\textrm{id}\mathcal{L}_\textrm{id}
\label{equ:rde}
\end{equation}

We infer the embeddings from decompressed images and original images via face alignment module $A$ and face recognition model $R$ with shared weights $\bm{\theta}$ as

\begin{equation}
\bm{e} = R_{\bm{\theta}} (A(\bm{x})) \textrm{\quad and\quad} \bm{\hat e} = R_{\bm{\theta}} (A(\bm{\hat x}))
\label{equ:align}
\end{equation}

Since we use the embeddings to match images with identities, if the cosine distance between $\bm{\hat e}$ and $\bm{e}$ is small enough, decompressed images are likely to maintain parity in recognition performance.

To simplify the hyperparameter fine-tuning, we set $\lambda_\textrm{id}$=1.0 or 0.0 and only fine-tuned $\lambda_\textrm{rate}$. IPR loss with $\lambda_\textrm{id}$=0.0 is equivalent to Reconstruction-only loss. Simultaneously fine-tuning $\lambda_\textrm{rate}$ and $\lambda_\textrm{id}$ may help the model get a better balance between multiple objectives, but that is not the focus of this work.

Distinguished from other task-specific or face-specific compression methods, our method considers to:

\begin{itemize}

\item{\textbf{Compress unaligned images}}\quad Although the recognition performance is only related to the aligned face region, our method compresses either unaligned or aligned images. It uses reconstruction loss to optimize the overall image quality, not only for face regions. Compressing unaligned images is for practical retention decisions to save unaligned images for reference, and allowance for changing the alignment methods.

\item{\textbf{Avoid fine-tuning downstream models}}\quad Our method does not (end-to-end) fine-tune or retrain the downstream model by any face recognition loss. First, this can avoid additional effort to fine-tune hyperparameters of the recognition loss and reduce the data scale and training time required for fine-tuning a high-complexity face recognition model. Second, this can decouple the compression and downstream models. We can use the same downstream model to achieve variable bit rates (training compression models with different hyperparameters). Third, this helps maintain the robustness to downstream model change. When end-to-end fine-tuning the recognition model, the recognition model will adapt to the compressed images, allowing more unexpected distortion from the compression model. The exceeded distortion may degrade the recognition performance when another recognition model (trained on raw face images) is attached but not retrained on compressed images.

\end{itemize}

\subsection{Face Alignment Module}\label{sec:align}
For the framework (Fig. \ref{fig:framework}), we attach the face alignment module and face recognition model to calculate the embeddings. For alignment, we use the MTCNN model from FaceNet \cite{Schroff2015FaceNetAU} to detect faces. It outputs the bounding boxes and 5-point landmarks for each face. We select the face that has the largest size and locates centrally.

To compare different alignment orders in chosen datasets, we will use CelebA-HQ and CelebA-HQ-align mentioned in Section \ref{sec:datasets}. We do post-alignment and pre-alignment on these datasets, respectively.
% By using pre-alignment, we will get much smaller original images to compress, which means reduced file size in practice. However, alignment is an irreversible operation, and we cannot change the alignment algorithm when reusing these compressed aligned images.
% In this work, we show both alignment orders to satisfy different requirements (reduce payload or store raw images), and in the experiments, we will investigate how the compression model will perform on aligned images compared with unaligned ones.

\begin{figure}
    \centering
    \begin{subfigure}[b]{0.16\textwidth}
    \includegraphics[width=\textwidth]{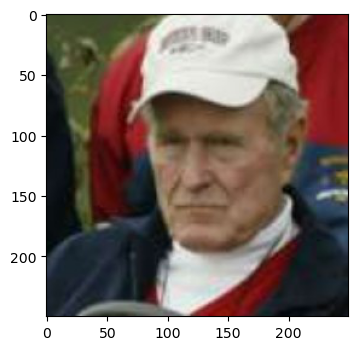}
    \end{subfigure}
    \begin{subfigure}[b]{0.16\textwidth}
    \includegraphics[width=\textwidth]{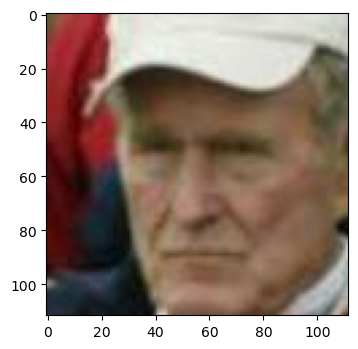}
    \end{subfigure}
    \begin{subfigure}[b]{0.15\textwidth}
    \includegraphics[width=\textwidth]{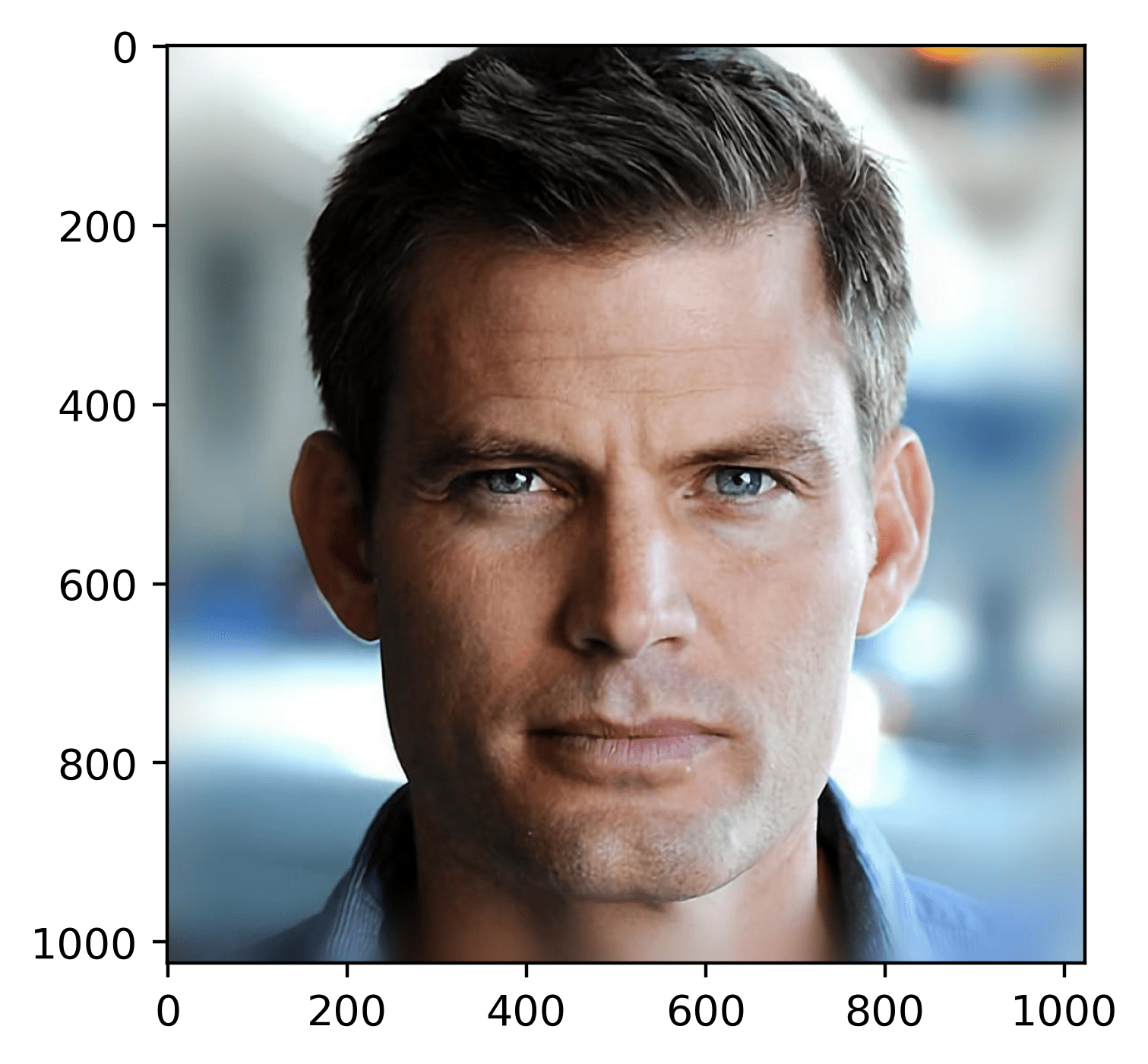}
    \end{subfigure}
    \begin{subfigure}[b]{0.15\textwidth}
    \includegraphics[width=\textwidth]{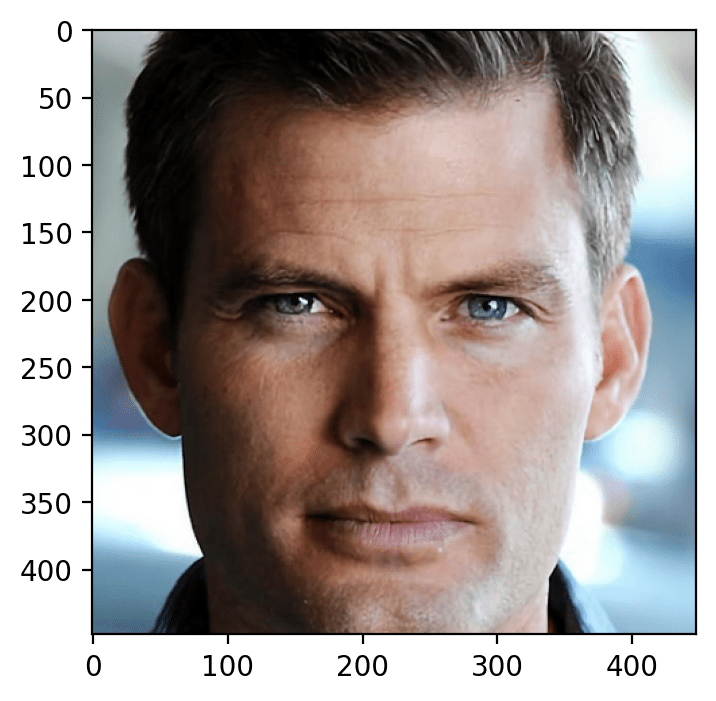}
    \end{subfigure}
    \begin{subfigure}[b]{0.15\textwidth}
    \includegraphics[width=\textwidth]{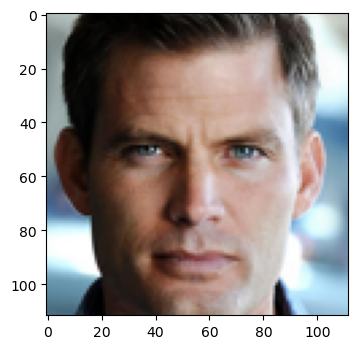}
    \end{subfigure}
    \caption{The results of alignment and resizing operation (from left to right). LFW images are aligned from 250x250 to 112x112. CelebA-HQ images are aligned from 1024x1024 to 448x448 and then further resized to 112x112.}
    \label{fig:align}
\end{figure}

Since our pre-trained model only accepts 112x112 images, we need to resize images for CelebA-HQ related high-resolution images after the alignment. Fig. \ref{fig:align} shows the additional resizing operation on aligned high-resolution face images, while the low-resolution face images only require an alignment for recognition.

%-------------------------------------------------------------------------

\section{Results}

\subsection{Datasets}\label{sec:datasets}

Table \ref{tab:dataset} shows the datasets for evaluating different image compression approaches. Images for all chosen datasets are available only in JPEG format; hence we use those as original images without compression. We use the LFW \cite{LFW} dataset as a low-resolution dataset, and CelebA-HQ \cite{karras2018progressive} dataset as a high-resolution dataset (We only use a subset of faces that are also used by CelebAMask-HQ \cite{lee2020maskgan}). Since CelebA-HQ dataset does not have an official train/test split, we create a split to make the number of train identities approximately equal to the number of test identities. We use the face alignment algorithm mentioned in Section \ref{sec:align} to align the images of CelebA-HQ and create CelebA-HQ-align. CelebA-HQ-align uses pre-alignment, and the image size is 448x448. We also use CLIC dataset \cite{clic} to train a general compression model for comparison.

\begin{table*}
\centering
\begin{tabular}{cccccc}
\toprule
     Dataset Name &	\#Train Images &	\#Train Indentites &	\#Test Images &	\#Test Identities &	Raw image size\\
\midrule
LFW &	9k &	4k &	4k &	2k &	250x250\\
CelebA-HQ &	20k	 & 3k &	10k &	3k &	1024x1024\\
CelebA-HQ-align	 & 20k	 &3k &	10k	 &3k &	448x448\\
CLIC & 626 & N/A & 60 & N/A & Unspecified
\\\bottomrule
\end{tabular}
\caption{Dataset Information. Training on CLIC images uses small patches from the original images since the original images are too large. Training on other datasets is based on the whole image.}
\label{tab:dataset}
\end{table*}

\subsection{Metrics}
To evaluate the performance of our image compression models, we will be utilizing the following metrics.
\subsubsection{Evaluating Compression Ratio}
We use the bits per pixel (BPP) to measure the performance of the compression ratio. The BPP is measured as
\begin{equation}
    \textrm{BPP} = \frac{\textrm{File size in bits}}{\text{\# pixels in one image}}
\end{equation}

We target low BPPs in this work (2x to 5x compression ratio of CRF-23-HEVC files).
% Our goal is to improve compression by 5x on File size and BPP (getting 1/5 File size and BPP) compared with HEVC files.

\subsubsection{Evaluating Recognition Performance}
For evaluating the accuracy of recognition tasks, we use FRR@FAR=1\% metric. The metrics are measured using the following process: %\pbcomment{I believe computing FRR/FAR is standardized. You can consider removing this section in interest of space.}
\begin{enumerate}
    \item Put $N$ images per identity into the gallery set and the rest into the query set.
    \item For each image in the query set, measure the embedding distance between this image and all images in the gallery set.
    \item Label the lowest embedding distance from the same identity as \emph{same distance}, and the lowest embedding distance from any other not-same identities as \emph{not-same distance} for identity matching. 
    \item Plot the ROC curve for all images in the query set and calculate FRR@FAR=1\% (false rejection rate when false acceptance rate equals to 1\%).
\end{enumerate}

For the LFW dataset, we set $N=1$ to evaluate the recognition performance. Our recognition model is trained to have high accuracy on LFW and is not fine-tuned on CelebA-HQ. Hence for CelebA-HQ and CelebA-HQ-align datasets, we set $N=3$ to have improved representation for each identity in the gallery set. 
Note that our evaluation protocol is more challenging compared to the standard LFW face verification protocol \cite{LFW}. Our protocol uses a more comprehensive negative pair set to compare each query against the entire gallery. In contrast, the standard protocol uses a random sampling process to generate a limited set of image pairs. Our primary goal is to maintain recognition parity with CRF-23-HEVC compressed images while achieving higher compression ratios.

\subsubsection{Evaluating Image Quality}    
We evaluate image quality in terms of Peak signal-to-noise ratio (PSNR) and Multi-scale structural similarity index measure (MS-SSIM) \cite{1292216} metrics. The goal is to achieve a similar image quality to HEVC compressed images w.r.t. PSNR and MS-SSIM metrics when using Reconstruction-only loss or IPR loss (IP loss does not optimize image quality).

\begin{figure*}
    \centering
    \includegraphics[width=\textwidth]{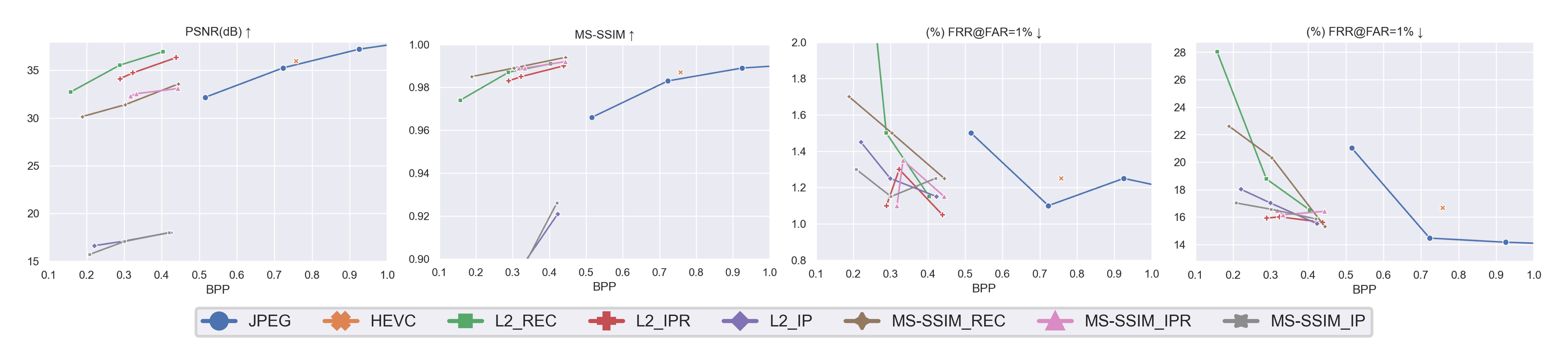}
    % \begin{subfigure}[b]{0.2\textwidth}
    % \includegraphics[width=\textwidth]{New_Figures/psnr_lfw.png}
    % \end{subfigure}
    % \begin{subfigure}[b]{0.2\textwidth}
    % \includegraphics[width=\textwidth]{New_Figures/msssim_lfw.png}
    % \end{subfigure}
    % \begin{subfigure}[b]{0.2\textwidth}
    % \includegraphics[width=\textwidth]{New_Figures/fr1_lfw.png}
    % \end{subfigure}
    % \begin{subfigure}[b]{0.2\textwidth}
    % \includegraphics[width=\textwidth]{New_Figures/fr01_lfw.png}
    % \end{subfigure}
    \caption{The image quality and recognition performance of JPEG, HEVC, and our method on the LFW dataset. From left to right: Image quality results (PSNR and MS-SSIM), recognition results with CurricularFace model, recognition results with CosFace model.}
    \label{fig:iq_r_lfw}
    % \caption{Image quality performance of JPEG, HEVC and our methods on LFW dataset (The higher points have better performance)}
    % \label{fig:iq_r_lfw}
\end{figure*}

\subsection{Experiment Settings}
To compare with standard compression techniques, we use \textbf{HEVC}, one of the latest state-of-the-art coding methods, with CRF=23 as our main baseline. We are focusing on HEVC intraframe compression performance on still images. The video compression performance is not evaluated since the embedding system for face recognition rarely transfers the whole video for latency consideration. The compression quality is set to balance the BPP and recognition performance trade-off. We also present the results for \textbf{JPEG} series. Since the original format is JPEG, we recompress the images at varying JPEG compression levels.

% We trained our models with the mentioned loss functions and name the models as follow:
We trained models with different combinations of reconstruction and identity preserving loss functions:
Reconstruction-only loss (equation \ref{equ:rd}, \textbf{L2\_REC} and \textbf{MS-SSIM\_REC}), IPR loss (equation \ref{equ:rde}, \textbf{L2\_IPR} and \textbf{MS-SSIM\_IPR}), IP loss (equation \ref{equ:re}, \textbf{L2\_IP} and \textbf{MS-SSIM\_IP}). The prefixes ``L2'' or ``MS-SSIM'' indicate the form of reconstruction loss functions (equation \ref{equ:d}) we use for training. For the models trained with IPR or IP loss function, the weights are initialized from those trained with Reconstruction-only loss.

We also train a general image compression model, where we use the same structure as our framework without face alignment and face recognition modules. The training images are the small patches from the original images (Table \ref{tab:dataset}), and we apply L2 loss as the reconstruction loss function. We name the model as \textbf{L2\_baseline}.

When training with a recognition model, we use a fixed CurricularFace\cite{huang2020curricularface} face recognition model (pre-trained on refined MS1M-v2 dataset \cite{deng2019arcface} and SOTA on IJB-C dataset \cite{maze2018iarpa}). During the evaluation, we use the CurricularFace model to compare the performance and a CosFace\cite{Wang2018CosFaceLM} (pre-trained on VGGFace2 dataset \cite{8373813}) model to evaluate the robustness of our framework on an unseen recognition model.

\subsection{Quantitative Results}

In the following sections, we discuss our quantitative results on LFW dataset (Fig. \ref{fig:iq_r_lfw}) and CelebA-HQ(-align) datasets (Fig. \ref{fig:iq_r_celeba}). We have the following key findings in our experiments. 

\subsubsection{IPR Loss Balances the Trade-off between Image Quality, Recognition Performance, and File Size} \label{high_quan}

In Fig. \ref{fig:iq_r_lfw}, We show image quality performance and recognition performance on models trained with Reconstruction-only Loss, IPR loss, and IP loss on the LFW dataset.

As the results show, models trained with Reconstruction-only loss (L2\_REC and MS-SSIM\_REC) mostly have better image quality performance than those trained with IPR or IP loss. For example, L2\_REC has 1.43 more PSNR than L2\_IPR, and 18.45 more PSNR than L2\_IP when BPP is about 0.3. However, L2\_REC and MS-SSIM\_REC have worse recognition performance (FRR@FAR=1\% increases to 4.16\% for L2\_REC) when the BPPs are less than 0.25.

Models trained with IPR loss (L2\_IPR and MS-SSIM\_IPR) improve recognition performance without significantly decreasing image quality. For instance, L2\_IPR has only 1.43 less PSNR but 0.4\% less FRR@FAR=1\% than L2\_REC when BPP is about 0.3. But training with IPR loss leads to slightly higher BPP. It moves curves to the right side in Fig. \ref{fig:iq_r_lfw}, and the lowest BPP is 0.29.

Models trained with IP loss show the path to lower BPPs with acceptable recognition performance. MS-SSIM\_IP (BPP=0.2) increases only 0.05\% FRR@FAR=1\% than HEVC, but gets 73\% less BPP than HEVC.  However, these models show a significant deterioration in image quality metrics. When BPP is about 0.3, L2\_IP has 18.45 less PSNR and 0.097 less MS-SSIM than L2\_REC, and MS-SSIM\_IP has 14.34 less PSNR and 0.102 less MS-SSIM than MS-SSIM\_REC.

We also investigate the trade-off between image quality performance, recognition performance, and BPP in Fig. \ref{fig:iq_r_lfw}: Models trained on IP loss or Reconstruction-only loss can reach low BPPs, which are less than 0.25. They can also get a better recognition performance or image quality than each other, respectively. L2\_IP has 0.25\% less FRR@FAR=1\% and 18.45 less PSNR than L2\_REC when BPP is about 0.3. When models are trained with IPR loss (L2\_IPR and MS-SSIM\_IPR), image quality and recognition performance are improved, but the resulting BPPs are larger. We find no models with IPR loss can reach BPP=0.25. We also observe that when the BPPs decrease below 0.25, the models with IP loss (purple and gray curves) have a sharp deterioration in FRR@FAR=1\% values. This may indicate a lower bound of BPP to maintain parity on recognition performance.
    
We summarize that one can focus on either image quality (L2\_REC and MS-SSIM\_REC) or recognition performance (L2\_IP and MS-SSIM\_IP), but not both when trying to have a BPP lower than 0.25. When the BPP is lower than 0.25 by focusing on recognition performance, models with IP loss will start to degrade in recognition performance.

\begin{figure*}[!htb]
    \centering
     \includegraphics[width=\textwidth]{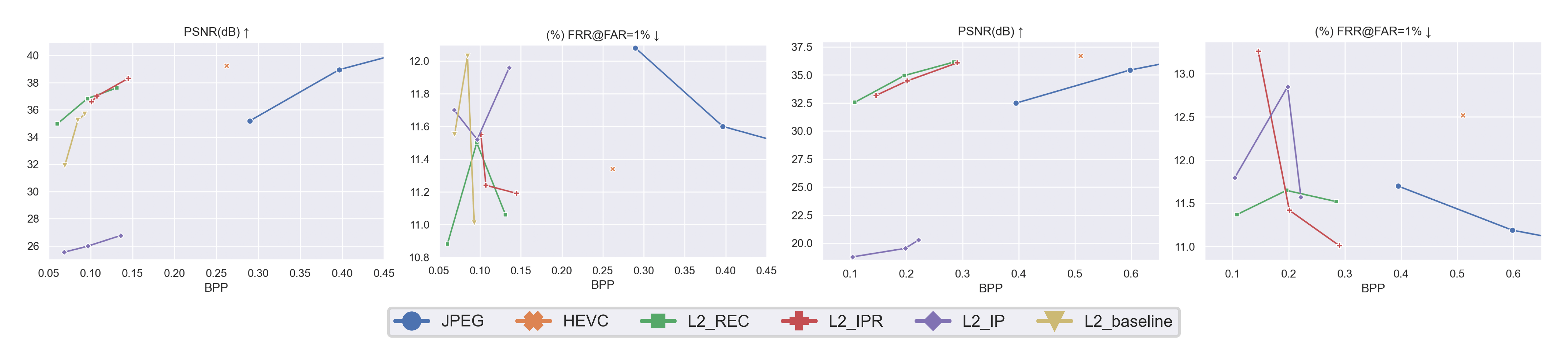}
    % \begin{subfigure}[b]{0.4\textwidth}
    % \includegraphics[width=\textwidth]{New_Figures/psnr_celeba.png}
    % \end{subfigure}
    % \begin{subfigure}[b]{0.4\textwidth}
    % \includegraphics[width=\textwidth]{New_Figures/psnr_celeba_align.png}
    % \end{subfigure}
    \caption{Image quality and recognition performance of JPEG, HEVC and our methods on CelebA-HQ (left two plots) and CelebA-HQ-align (right two plots) with CurricularFace model} 
    \label{fig:iq_r_celeba}
\end{figure*}

\textbf{Discussion on performance trends}\quad In Fig.\ref{fig:iq_r_celeba}, we observe that for high-resolution datasets (CelebA-HQ and CelebA-HQ-align), the performance trend is unstable. Lower BPP can get better performance. The advantage of IPR loss is less significant than that of the LFW dataset. This may be due to the poor generalization of the pre-trained recognition model on these datasets. The distribution histogram of the recognition embedding distance shows the recognition performance on original images is poor: There are a lot of same-identity pairs with high distance values, indicating that the recognition model cannot correctly match these images with the identities. This error could introduce noise as the IPR loss uses this feedback from the recognition model to generate decompressed images.

Another possible reason might be the resizing operation in the face alignment module on the high-resolution datasets. The face alignment module for CelebA-HQ(-aligned) images resizes aligned images from (448, 448) to (112, 112) to fit the recognition model’s input resolution (as shown in Fig. 5). In this case, IPR loss is more likely to ignore artifacts on the face region (also discussed in Section \ref{high_qua}).

In Fig. \ref{fig:iq_r_lfw}, for the low-resolution dataset (LFW), thanks to avoiding both reasons listed above, the noise becomes much smaller but still exists (for L2\_IPR and MS-SSIM\_IPR curves). This may be caused by the noise from balancing the tradeoff between image quality and recognition performance.

\subsubsection{Training against Face Domain Enables Higher Compression Ratio}

    To compare the performance between general compression models and domain-specific compression models, we trained the general compression model on the CLIC training set. We evaluated it on the CelebA-HQ test set (Fig. \ref{fig:iq_r_celeba}). Both datasets have similar resolutions to reduce the influence of scale change. The results show that our models trained on the CelebA-HQ dataset (L2\_REC) have better image quality performance and recognition accuracy. For example, when BPP is about 0.06, L2\_REC has 3.08 more PSNR and 0.67\% less FRR@FAR=1\% than L2\_baseline. Comparing models with about 35.50 PSNR, L2\_REC has 29\% less BPP and 1.15\% less FRR@FAR=1\% than L2\_baseline. This shows that when we focus on the face domain, compression models trained against face images benefit the compression ratio without decreasing image quality and recognition accuracy.
    
    We also find a large gap between expected BPPs and actual BPPs of the general compression model on face datasets. In Fig. \ref{fig:iq_r_celeba}, for the leftmost point of the L2\_baseline curve, when we set the target BPP to be 0.100, the validation BPP on the CLIC test set is 0.095, which is close to the target, but the test BPP on CelebA-HQ test set is 0.069, which shows a non-trivial decrease on the BPP. For this unexpected low BPP, the compressed images have a low PSNR score of 31.89. The drop of BPPs may be caused by the domain gap between CLIC and CelebA-HQ. It increases the uncertainty in compression ratio and image quality if the general model is applied to a new dataset on specific domains and will be an unstable factor in transmission. %\pbcomment{Has been rephased. Please read again and see if that makes sense.}

\subsubsection{Robustness to Downstream Model Change}

In Fig. \ref{fig:iq_r_lfw}, we show the recognition performance of our method when training on CurricularFace but evaluating on another recognition model CosFace \cite{Wang2018CosFaceLM}. In the results, MS-SSIM\_IP (BPP=0.3) shows 0.11\% less FRR@FAR=1\% and 60\% less BPP than HEVC result. L2\_IPR (BPP=0.29) shows 0.76\% less FRR@FAR=1\% and 62\% less BPP than HEVC result. The results with IPR loss or IP loss show that the compression model trained with one recognition model can have good generalization performance with another recognition model on the LFW dataset.
    
We have shown that models with IP loss (L2\_IP and MS-SSIM\_IP) have significant distortion (shown in decreased PSNR in Fig. \ref{fig:iq_r_lfw}) on the images when the quality level is set to the lowest (leftmost points of the curves). But still, the models perform at most 0.35\% more FRR@FAR=1\% than HEVC baseline with the CurricularFace model. This implies that IP loss corrupts only task-unrelated regions but maintains recognizable face features. The result with the CosFace model also depicts that good generalization performance may benefit from these maintained face features.

\subsection{Qualitative Results} \label{vis}
\begin{figure}[!htb]
\captionsetup[subfigure]{labelformat=empty}
    \centering
    \begin{subfigure}[b]{0.15\textwidth}
    \includegraphics[width=\textwidth]{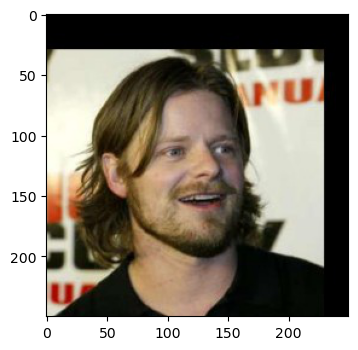}
    \caption{\scriptsize Original\\BPP=9.76}
    \end{subfigure}
    \begin{subfigure}[b]{0.15\textwidth}
    \includegraphics[width=\textwidth]{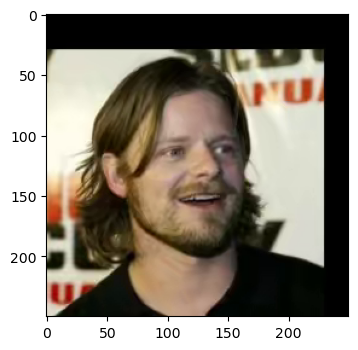}
    \caption{\scriptsize HEVC\\BPP=0.76}
    \end{subfigure}
    \begin{subfigure}[b]{0.15\textwidth}
    \includegraphics[width=\textwidth]{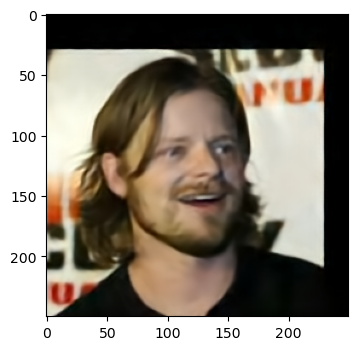}
    \caption{\scriptsize L2\_REC\\BPP=0.16}
    \end{subfigure}
    \begin{subfigure}[b]{0.15\textwidth}
    \includegraphics[width=\textwidth]{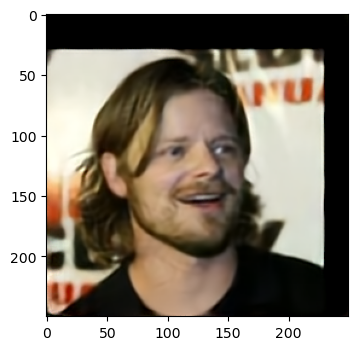}
    \caption{\scriptsize MS-SSIM\_REC\\BPP=0.19}
    \end{subfigure}
    \begin{subfigure}[b]{0.15\textwidth}
    \includegraphics[width=\textwidth]{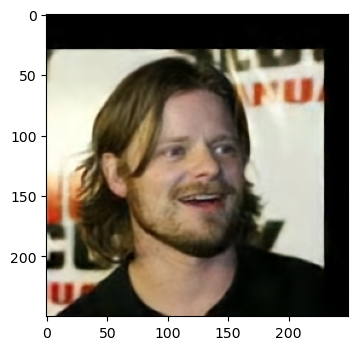}
    \caption{\scriptsize L2\_IPR\\ BPP=0.29}
    \end{subfigure}
    \begin{subfigure}[b]{0.15\textwidth}
    \includegraphics[width=\textwidth]{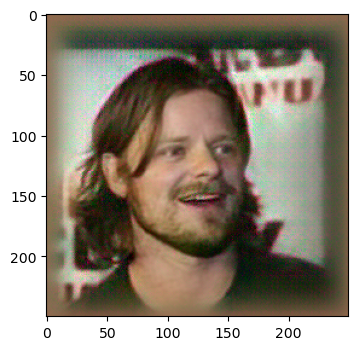}
    \caption{\scriptsize L2\_IP\\BPP=0.22}
    \end{subfigure}
    % \begin{subfigure}[b]{0.15\textwidth}
    % \includegraphics[width=\textwidth]{Visual/lfw/CM_MS-SSIM_REC_EMB_low.png}
    % \caption{\scriptsize MS-SSIM\_IPR\\BPP=0.32}
    % \end{subfigure}
    % \begin{subfigure}[b]{0.15\textwidth}
    % \includegraphics[width=\textwidth]{Visual/lfw/CM_MS-SSIM_EMB_low.png}
    % \caption{\scriptsize MS-SSIM\_IP\\BPP=0.21}
    % \end{subfigure}
    \caption{Visualization results on LFW datasets}
    \label{fig:vis_lfw}
\end{figure}

\begin{figure}[!htb]
\captionsetup[subfigure]{labelformat=empty}
    \centering
    \begin{subfigure}[b]{0.15\textwidth}
    \includegraphics[width=\textwidth]{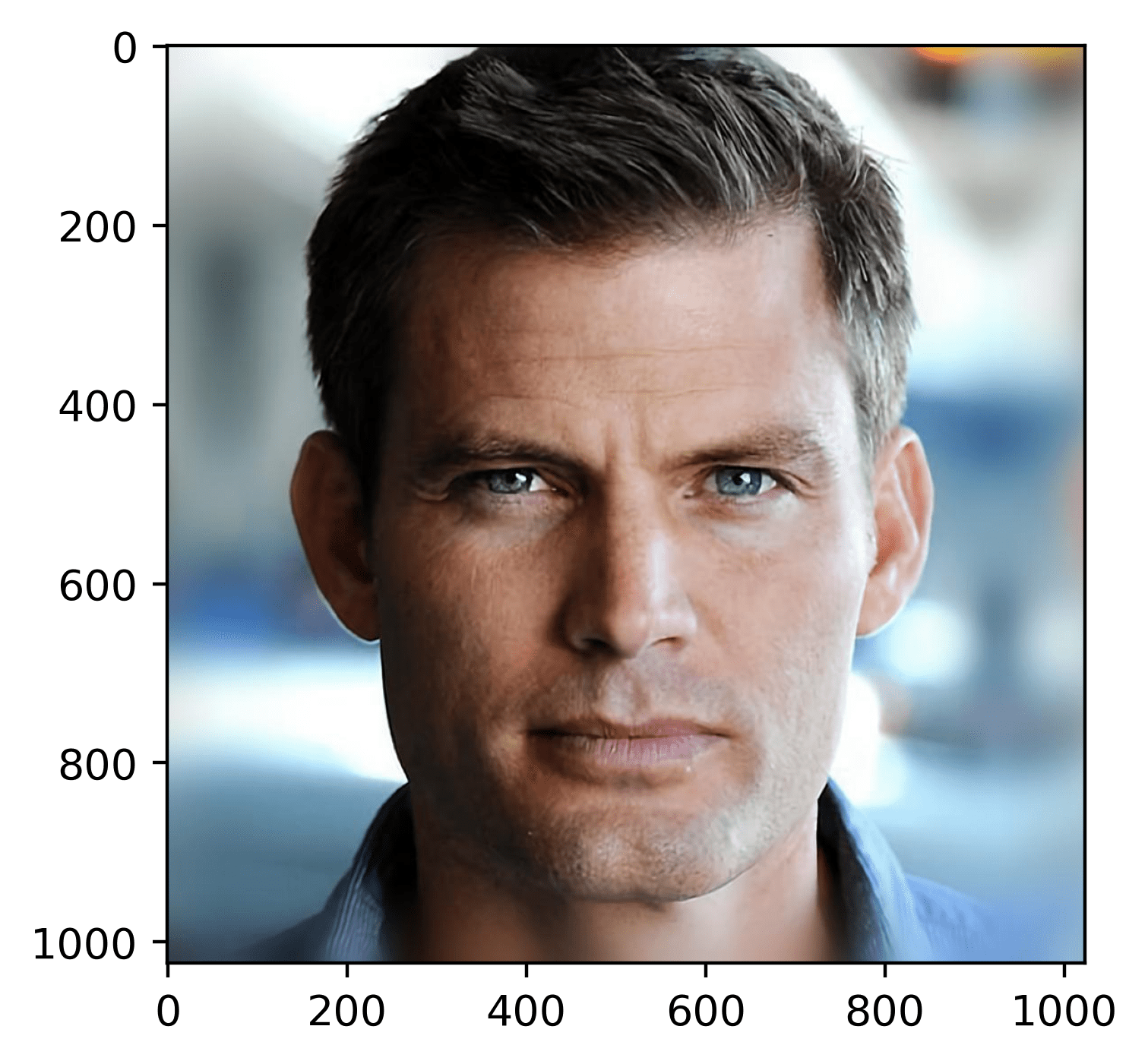}
    \caption{\scriptsize Original\\BPP=6.25}
    \end{subfigure}
    % \begin{subfigure}[b]{0.16\textwidth}
    % \includegraphics[width=\textwidth]{Visual/celeba/HEVC.png}
    % \caption{HEVC\\BPP=0.26}
    % \end{subfigure}
    \begin{subfigure}[b]{0.15\textwidth}
    \includegraphics[width=\textwidth]{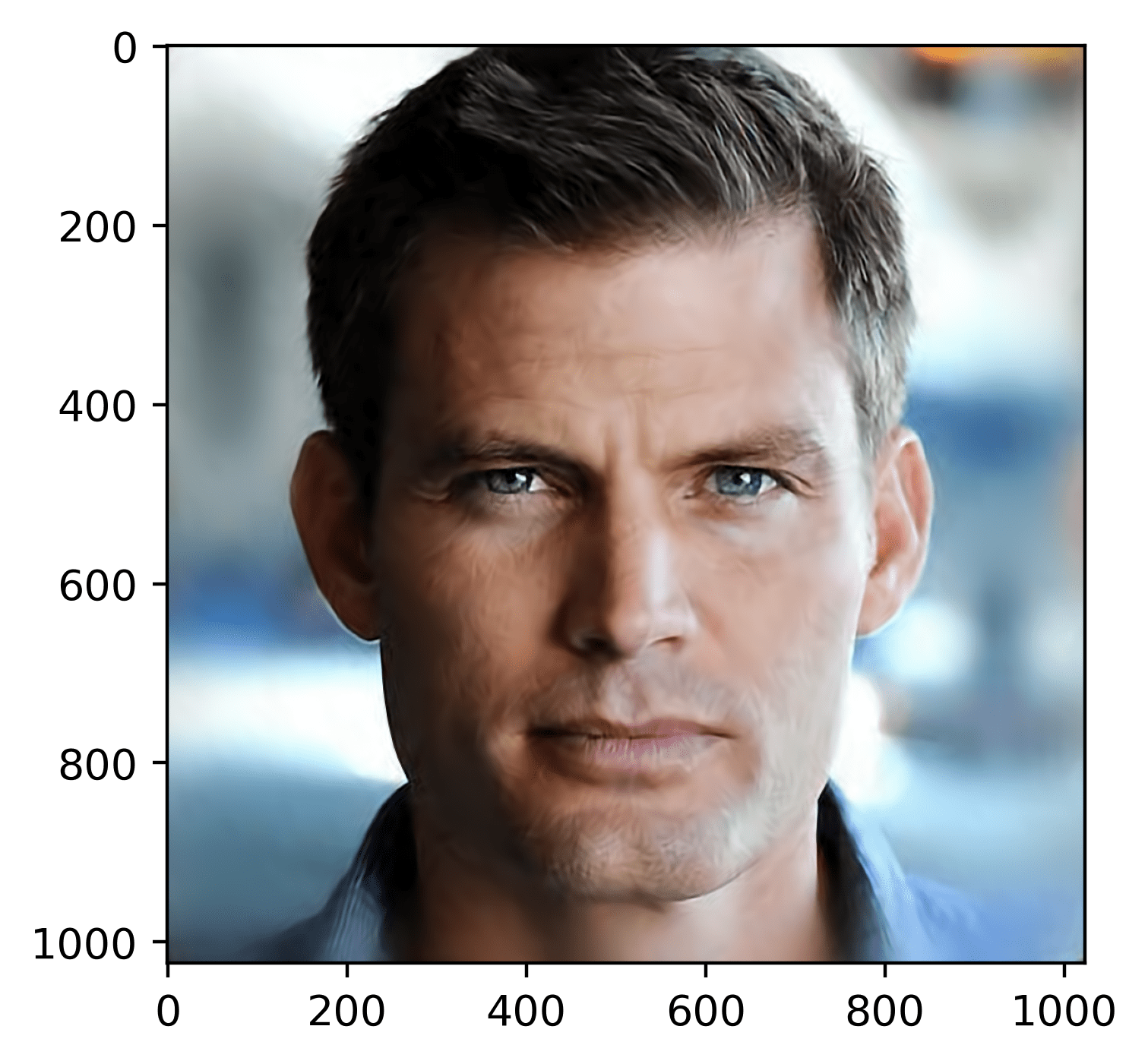}
    \caption{\scriptsize L2\_REC\\BPP=0.06}
    \end{subfigure}
    \begin{subfigure}[b]{0.15\textwidth}
    \includegraphics[width=\textwidth]{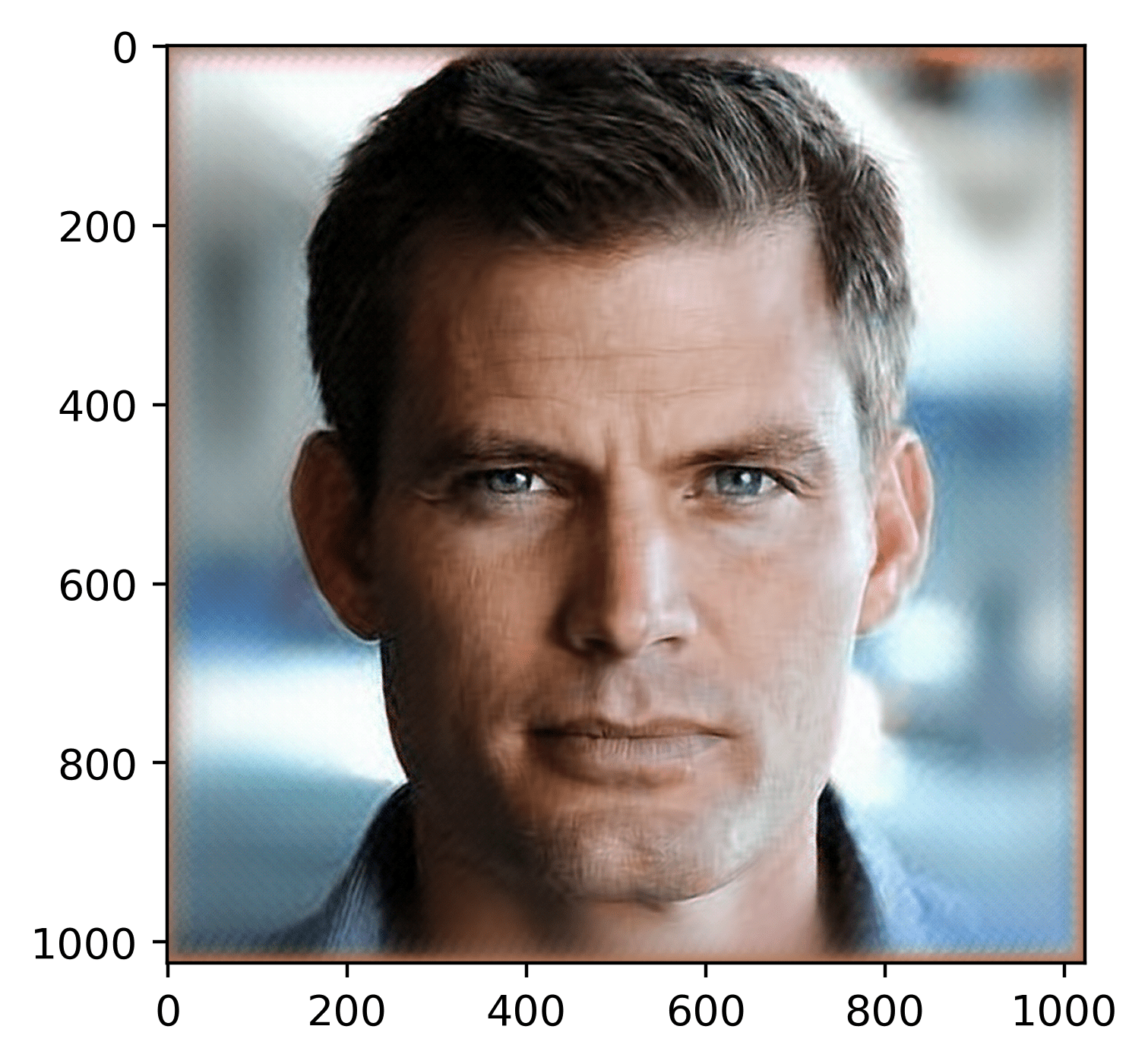}
    \caption{\scriptsize L2\_IP\\BPP=0.07}
    \end{subfigure}
    \centering
    \begin{subfigure}[b]{0.15\textwidth}
    \includegraphics[width=\textwidth]{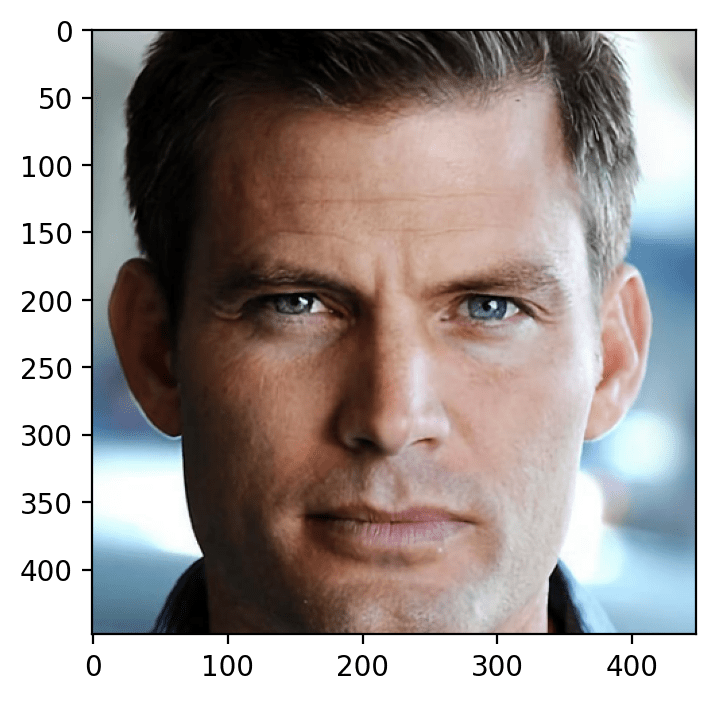}
    \caption{\scriptsize Original\\BPP=9.69}
    \end{subfigure}
    % \begin{subfigure}[b]{0.15\textwidth}
    % \includegraphics[width=\textwidth]{Visual/celeba_align/HEVC.png}
    % \caption{HEVC\\BPP=0.51}
    % \end{subfigure}
    \begin{subfigure}[b]{0.15\textwidth}
    \includegraphics[width=\textwidth]{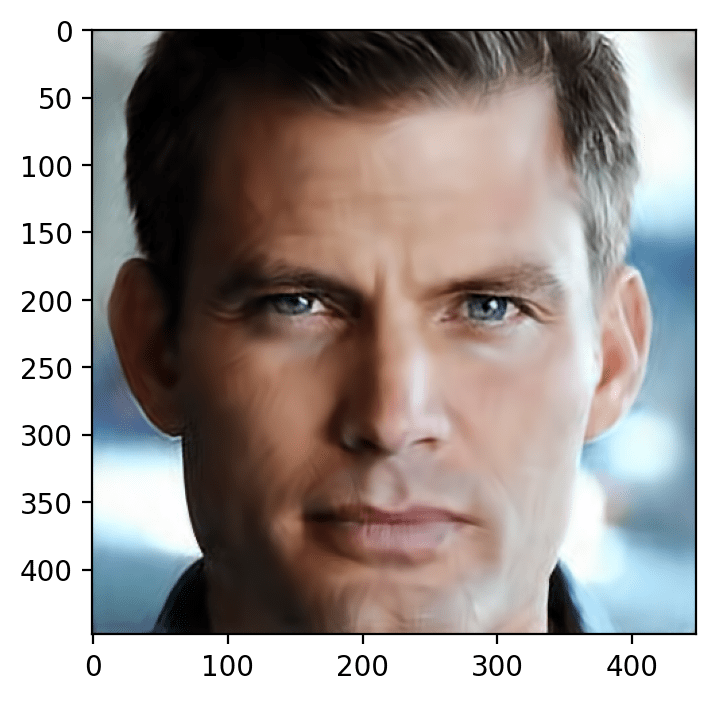}
    \caption{\scriptsize L2\_REC\\BPP=0.11}
    \end{subfigure}
    \begin{subfigure}[b]{0.15\textwidth}
    \includegraphics[width=\textwidth]{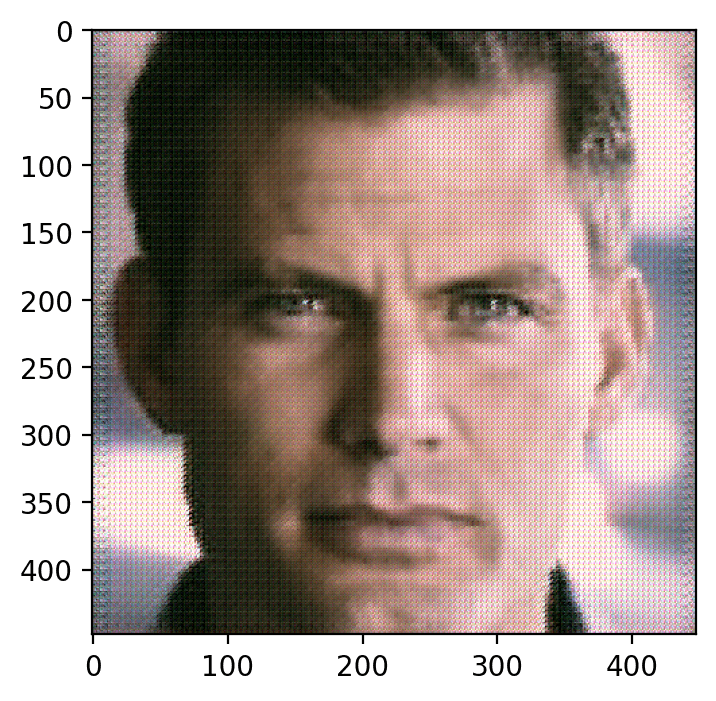}
    \caption{\scriptsize L2\_IP\\BPP=0.10}
    \end{subfigure}
    \caption{Visualization results on CelebA-HQ (Upper) and CelebA-HQ-align (Lower) datasets}
    \label{fig:vis_celeba}
\end{figure}

In this section, we present qualitative visualizations imagery from LFW and CelebA-HQ(-align) datasets. Fig. \ref{fig:vis_lfw}-\ref{fig:vis_celeba} show the reconstructed images with typical compression methods and qualities. All the images use the lowest quality level (the leftmost points of curves) to better visualize possible artifacts. We make the following observations: 

% \subsubsection{Visual Artifact Analysis}

%     In LFW dataset (Fig. \ref{fig:vis_lfw}), models with Reconstruction-only loss (L2\_REC and MS-SSIM\_REC) have blur artifacts on the face region. MS-SSIM\_REC has less notable distortion (on eyes and mustache) than L2\_REC, which shows the advantage of MS-SSIM loss. Model with IPR loss (L2\_IPR) produces clear face details (eyes and For models with IP loss (L2\_IP), the boundary of the image is corrupted and the background becomes darker.
    
%     In CelebA-HQ dataset (Fig. \ref{fig:vis_celeba}), models with Recon-struction-only loss (L2\_REC) have some striped texture on the skin. The boundary of models with IP loss (L2\_IP) is corrupted and have obvious artifacts on the skin and background. In CelebA-HQ-align dataset  (Fig. \ref{fig:vis_celeba}), the images of L2\_IP turn red and have obvious artifacts.

    \subsubsection{IPR loss vs. IP loss} %IPR loss vs. IP loss

    For models with IPR loss, the results of L2\_IPR and  MS-SSIM\_IPR do not show notable artifacts on the images. However, models with IP loss (L2\_IP and MS-SSIM\_IP) have strong distortion on the images. The difference between IPR loss and IP loss is the reconstruction loss term $\mathcal{L}_\textrm{rec}$. Models with IPR loss optimize reconstruction loss to maintain high image quality, but models with IP loss allow more distortion if the predicted embedding $\bm{\hat e}$ is close enough to the original embedding $\bm e$.
    
    \textbf{Discussion on artifacts from IP loss}\quad Models with IP loss show different artifacts on unaligned datasets (LFW and CelebA-HQ) and the aligned dataset (CelebA-HQ-align). The unaligned compressed images show the corrupted boundary and distorted background. At the same time, in aligned datasets (CelebA-HQ-align), the visualized results have mesh artifacts on the whole image, where the face region is strongly distorted. The influence of the face alignment module on unaligned images is shown: Face alignment will discard the boundary and only crop the center face region of the images. Therefore, it allows the compression model with IP loss to focus on the center face region and reduces bitrate for the boundary and background.

    % Hence, we conclude that on unaligned images, the models with IP loss will corrupt the boundary and background. On aligned images, the models with IP loss will globally redden and corrupt the images. But both of them do not lead to any significant deterioration on recognition performance. The models trained with IPR loss can reduce the corruption of the background and boundary on unaligned images or the global distortion on aligned images produced by IP loss.

    \textbf{Discussion with mask-based compression}\quad The result of IP loss is visually similar to mask-based compression methods \cite{8943263,1017274} on LFW and CelebA-HQ datasets, which explicitly or implicitly mask out the background region and allocate more bits on the ROI to maintain its parity. These methods require explicit ROI annotations, which have more annotation costs than our method. There can be a mismatch between ROIs expected by humans and the feature regions that downstream models are using. Our method avoids this mismatch by directly using the feedback from downstream models to adjust the allocation of bits.
    
    \subsubsection{High-resolution vs. Low-resolution}\label{high_qua}

    In the LFW low-resolution dataset, L2\_IPR and L2\_IP are shown to produce sharp and clear face regions. In CelebA-HQ and CelebA-HQ-align high-resolution datasets, all models have blurring artifacts and striped texture on the face region when the BPPs are low than 0.075 (CelebA-HQ) and 0.15 (CelebA-HQ-align). Even though the face region is not so clear, the recognition accuracy is maintained: L2\_REC has 0.46\% less FRR@FAR=1\% for CelebA-HQ and 1.15\% less FRR@FAR=1\% for CelebA-HQ-align than HEVC baseline.

    The influence of the resizing operation on high-resolution datasets has been discussed in \ref{high_quan}, and the visualization results support the point. The pre-trained recognition model requires CelebA-HQ images to be resized, as shown in Fig. \ref{fig:align}. Therefore, some artifacts like blurring or striped texture will be invisible to the recognition model. After resizing, the vanishing of artifacts can make the IPR or IP loss ignore these artifacts.

    Overall, we show that resizing mitigates the influence of artifacts on recognition performance. Therefore, IPR loss and IP loss are less likely to maintain the details in the face region on high-resolution datasets.

%-------------------------------------------------------------------------

\section{Conclusion}

In this work, we presented an end-to-end image compression framework trained to leverage domain-specific features to achieve higher compression ratios while maintaining accuracy on downstream tasks. We presented an Identity Preserving Reconstruction (IPR) loss for the face recognition domain, which achieves BPP values $\sim38\%$ and $\sim42\%$ of CRF-23 HEVC compression for LFW and CelebA-HQ datasets, respectively, while maintaining parity in recognition accuracy. Our approach does not require fine-tuning the downstream task model (face recognition model). We also evaluated our approach to unseen downstream recognition models, highlighting the generalizability and robustness of the compressed imagery generated by our proposed framework. In conclusion, our work shows that the proposed framework with IPR loss can guide compression models to improve compression ratios by focusing on essential image features for downstream task accuracy.

{\small
\bibliographystyle{ieee_fullname}
\bibliography{egbib}
}

\end{document}